\documentclass[letterpaper,10 pt,conference]{ieeeconf}

\usepackage{graphicx}
\usepackage{amsmath}
\usepackage[font={footnotesize},labelfont=bf]{subcaption}
\usepackage{cite}
\usepackage{multirow}
\usepackage{booktabs}
\usepackage[detect-weight]{siunitx}
\usepackage[printonlyused, nolist]{acronym}
\let\labelindent\relax\usepackage[inline]{enumitem}
\usepackage{pifont}
\usepackage[activate={true,nocompatibility},final,tracking=true,kerning=true,spacing=true,factor=1100,stretch=20,shrink=20]{microtype}
\usepackage[nameinlink]{cleveref}
\usepackage[shortcuts]{extdash}
\usepackage[dvipsnames]{xcolor}
\usepackage{balance}

\usepackage{tikz}
\usepackage{textcomp}
\usepackage{lipsum}

\IEEEoverridecommandlockouts
\newcommand{\cmark}{\ding{51}}
\newcommand{\xmark}{\ding{55}}
\newcommand{\etal}{\textit{et al.}}
\graphicspath{{images/}{diagrams/}}
\begin{acronym}[CNNs] 
\acro{NIR}{Near Infra-Red}
\acro{FIR}{Far Infra-Red}
\acro{ROI}{Region of Interest}
\acro{RGB}{RGB}
\acro{IR}{Infrared}
\acro{LWIR}{Longwave Infrared}
\acro{MI}{Mutual Information}
\acro{MSE}{Mean Squared Error}
\acro{DASC}{Dense Adaptive Self-Correlation}
\acro{LSS}{Local Self-Similarity}
\acro{CNN}{Convolutional Neural Network}
\acrodefplural{CNN}[CNNs]{Convolutional Neural Networks}
\acro{IoU}{Intersection over Union}
\acro{mIoU}{mean Intersection over Union}
\end{acronym}

\newlist{contributions}{enumerate}{3}
\setlist[contributions,1]{label=\textbf{\arabic*)}, ref=\textbf{(\arabic*)}}
\creflabelformat{contributionsi}{#2#1#3}
\crefname{contributionsi}{contribution}{contributions}
\Crefname{contributionsi}{Contribution}{Contributions}
\newlist{reasons}{enumerate*}{3}
\setlist[reasons,1]{label={(\arabic*)}, ref={(\arabic*)}}
\creflabelformat{reasonsi}{#2#1#3}
\crefname{reasonsi}{}{}
\Crefname{reasonsi}{}{}
\definecolor{g}{RGB}{150, 150, 150}

\title{\LARGE\bf{}There and Back Again: Self-supervised\\Multispectral Correspondence Estimation}
\author{Celyn Walters$^{1}$, Oscar Mendez$^{1}$, Mark Johnson, Richard Bowden$^{1}$%
	\thanks{$^{1}$Celyn Walters \etal~with Centre for Vision, Speech and Signal Processing (CVSSP), University of Surrey, UK
	{\tt\small c.walters@surrey.ac.uk}}
}
\newcommand\copyrighttext{%
	\footnotesize This work has been submitted to the IEEE for possible publication.
	Copyright may be transferred without notice, after which this version may no longer be accessible.
}
\newcommand\copyrightnotice{%
	\begin{tikzpicture}[remember picture,overlay]
	\node[anchor=south,yshift=10pt] at (current page.south) {\fbox{\parbox{\dimexpr\textwidth-\fboxsep-\fboxrule\relax}{\copyrighttext}}};
	\end{tikzpicture}%
}
\begin{document}
\maketitle
\copyrightnotice
\thispagestyle{empty}
\pagestyle{empty}

\begin{abstract}
Across a wide range of applications, from autonomous vehicles to medical imaging, multi-spectral images provide an opportunity to extract additional information not present in color images.
One of the most important steps in making this information readily available is the accurate estimation of dense correspondences between different spectra.

Due to the nature of cross-spectral images, most correspondence solving techniques for the visual domain are simply not applicable.
Furthermore, most cross-spectral techniques utilize spectra-specific characteristics to perform the alignment.
In this work, we aim to address the dense correspondence estimation problem in a way that generalizes to more than one spectrum.
We do this by introducing a novel cycle-consistency metric that allows us to self-supervise.
This, combined with our spectra-agnostic loss functions, allows us to train the same network across multiple spectra.

We demonstrate our approach on the challenging task of dense RGB-\acs{FIR} correspondence estimation.
We also show the performance of our unmodified network on the cases of RGB-\acs{NIR} and RGB-RGB, where we achieve higher accuracy than similar self-supervised approaches.
Our work shows that cross-spectral correspondence estimation can be solved in a common framework that learns to generalize alignment across spectra.
\end{abstract}

\section{INTRODUCTION}
\label{sec:intro}
Solving the correspondence problem between two images is a fundamental problem in computer vision.
Its applications are widespread, including 3D reconstruction~\cite{Hartley2004}, motion estimation~\cite{Brady1995} and image registration~\cite{Lowe2004}.
Correspondence estimation from RGB to RGB is well understood, with many solutions using correlation~\cite{Lucas1981}, optimisation~\cite{Besl1992a}, hand-crafted feature descriptors~\cite{Lowe2004,Liu2015,Tola2010} or machine learning~\cite{Dosovitskiy2015a,Ilg2017,Sun2018a,Meister2018}.
However, in some use-cases, relying on the visible spectrum alone is insufficient.
For example, in areas such as autonomous navigation and visual surveillance, approaches using RGB cameras often fail at night, in poor weather, or due to extreme variability in lighting.
Using alternative spectra, such as \ac{IR}, is a commonly used technique to address these concerns.
\ac{NIR}, considering its similarities to RGB, may utilize vision algorithms developed for the visible spectrum.
\ac{FIR} gives much stronger thermal cues but requires specialized techniques.
Bridging this gap will lead to more capable applications.

Thermal sensors tend to be low resolution and lack fine detail at the far range, and high resolution thermal sensors are expensive.
Multispectral fusion can overcome the deficiencies of each individual sensor by combining their complementary properties.
However, most traditional applications require the images to be registered.

Relative to RGB, cross-spectral correspondence estimation approaches are scarce in the literature.
This is partially due to the availability and cost of sensors, in addition to the complications of solving the correspondence problem when photometric consistency between sensors does not hold true.
Most multispectral datasets/methods purposefully avoid tackling stereo disparity, by either
focusing on scenes at long range, where disparity is assumed to be negligible~\cite{Liu2019},
or by using a beam splitter to ensure coaxial camera centres~\cite{Hwang2015,Okazawa2019,Gebhardt2019}.
Unfortunately both approaches ignore possible stereo cues.
We argue that there is benefit in correctly modelling disparity as it provides more accurate sensor fusion while providing additional stereo cues.
Furthermore, neither high-resolution thermal cameras nor beam splitters are commodity items, and are therefore unlikely to be incorporated into consumer products.

Perhaps one of the most limiting factors behind research into multispectral correspondence is the poor availability of datasets.
As cross-spectral research is relatively unexplored, few annotated datasets are available and ground truth annotation is time-consuming and expensive.
Automatic annotation is an unsolved problem, and requires expensive precision equipment such as a laser scanner.
Due to unfamiliarity with the appearance differences, human annotators find annotation difficult.
Therefore, most provide sparse point annotation or weak labels such as bounding boxes~\cite{Hwang2015,Davis2005,Zhang2015d,Li2016a,Li2018}.
This motivates our proposed self-supervised approach which learns how to recover a dense flow field between RGB and \ac{IR}, and vice-versa. We make the following contributions:

\begin{contributions}
\item \textbf{Dense flow fields between spectra:}
\label{contribution:flow}
We present a spectrum-agnostic method to obtain cross-spectral flow fields at full image resolution as a way to solve the correspondence problem between different spectra.

\item \textbf{Self-supervised training:}
\label{contribution:selfsupervised}
We use a dual-spectrum siamese-like structure, utilizing cycle-consistency to avoid the need for ground truth correspondence.
This provides scalability, allowing us to significantly increase the training data seen by the system.

\item \textbf{Application to RGB-\ac{FIR}:}
We demonstrate RGB-\ac{FIR} correspondence estimation, which is seldom tackled.

\item \textbf{Application to RGB-RGB and RGB-\ac{NIR}:}
We further demonstrate our approach with competitive results for both RGB-\ac{NIR} and RGB-RGB correspondence.
\end{contributions}

\section{RELATED WORK}
\label{sec:lit}
Traditional solutions to the correspondence problem are to solve it sparsely, using optimisation, correlation, or feature-matching.
However, most approaches rely upon photometric consistency, and this assumption does not hold well for cross-spectral matching.
For wide-baselines, feature-matching approaches using hand-crafted feature descriptors are common (e.g. SIFT~\cite{Lowe2004}).
Although other features have been used for dense matching~\cite{Liu2015,Tola2010}, most approaches rely on features optimized for RGB and as a result, performance is lower when applied to other spectra~\cite{Ricaurte2014}.
Work that has attempted to adapt traditional techniques to RGB-\ac{NIR}~\cite{Brown2011,Saleem2014} tends to have drawbacks, e.g.~they scale poorly when applied densely over an entire image.
Kim \etal~proposed DASC~\cite{Kim2015a}, a dense descriptor which finds illumination/sharpness differences between both RGB and \ac{NIR} pairs.
Considering the closeness of \ac{NIR} to the visible spectrum and its similarities with greyscale, \ac{NIR}-RGB correspondence is perhaps easier than for other spectra like \ac{FIR}.
RGB-\ac{NIR}-specific approaches do not transfer well to RGB-\ac{FIR}.

Same-spectrum correspondence techniques can be adapted from RGB to other spectra.
For instance, thermal depth images~\cite{Hajebi2008} and thermal stereo odometry~\cite{Mouats2015}.
For RGB-\ac{FIR}, Li and Stevenson use a straight line matching scheme to \textit{register} stereo images~\cite{Li2017}, but a reliance on straight edges is likely to fail in less structured scenes.
By contrast, our approach uses feature losses to ensure we are robust to cross-spectral images and less structured scenes.

\ac{MI} does not rely upon photometric consistency, and is widely used for multispectral matching~\cite{Viola1997}.
However, \ac{MI} is not differentiable and cannot be used to supervise a neural network.
While \ac{MI} can be approximated (e.g.~MINE~\cite{Belghazi2018}), training a dense correspondence network using this approximation is inefficient and time consuming.
Our network structure, combined with feature losses and cycle-consistency, is the best-of-both-worlds, being both spectrum-agnostic and computationally efficient.

\subsection{Multispectral datasets}
\label{subsec:datasets}
\begin{figure*}[t]\centering
	\includegraphics[width=0.6\linewidth]{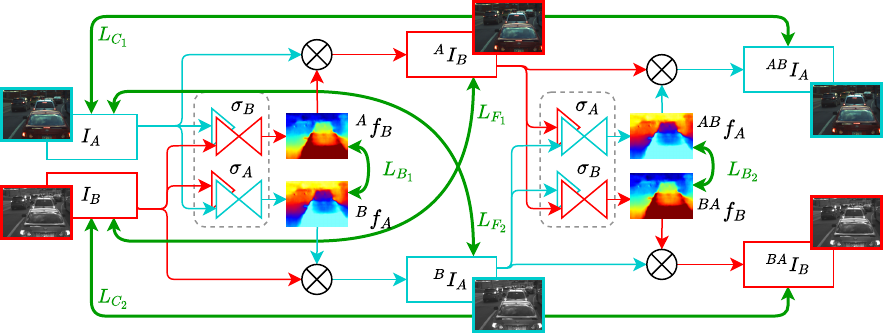}
	\caption{Network overview.
		Red borders and arrows relate \(I_A\)'s point of view, blue relates to \(I_B\)'s.
		Green lines represent different losses (labeled).
		The \(\bigotimes\) operation represents warping an image \(I\) with a flow field \(f\).
		Networks \(\sigma_A\) and \(\sigma_B\) relate to \cref{fig:module}; the same architecture but differently trained weights.}
	\label{fig:overview}
\end{figure*}
A current limitation for training and evaluating multispectral stereo correspondence techniques is the lack of suitable datasets.
The majority of RGB-\ac{FIR} datasets consist of pre-aligned image pairs, and therefore do not exhibit disparity.
For distant scenes, which approximate orthographic projection, alignment can be made with a simple homography registration.
Alternatively, images can be recorded with the same optical axis through the use of a beam-splitter.
This removes the need for alignment/correspondence estimation and applies to RGB/\ac{NIR} datasets EFPL~\cite{Brown2011}, RANUS~\cite{Choe2018}, and also RGB/\ac{FIR} datasets KAIST~\cite{Hwang2015}, Coaxials~\cite{Okazawa2019}, CAMEL~\cite{Gebhardt2019}.

Obtaining ground truth annotations for unaligned datasets is challenging, and human annotators rely on their familiarity with the visible spectrum.
As a result, it is not always possible to distinguish objects, particularly in the case of \ac{FIR}.
The KAIST~\cite{Hwang2015}, LITIV~\cite{Bilodeau2014} and PittsStereo~\cite{Zhi2018} datasets have weak bounding box annotations or very sparse point correspondences only.
The VAP dataset~\cite{Palmero2016} provides synchronized RGB, \ac{FIR} and depth images, with pedestrian segmentation in each modality.
To our knowledge, the CATS dataset~\cite{Treible2017} is the only with dense RGB-\ac{FIR} ground truth correspondence.
However, the annotations are poorly registered, making accurate quantitative evaluation impossible.

SODA from Li \etal~uses image-to-image translation to synthesize \ac{FIR} images from RGB, enabling the use of existing semantic labels~\cite{Li2019}.
However, these are limited to \ac{NIR} only and exhibit artefacts.
Image-to-image translation with \ac{FIR} is more difficult, and it is impractical to train a correspondence network on the output.
For example, a hot or cold car may have the same appearance in the visible spectrum.
Given an RGB image, it is not always possible to assign a correct temperature using image-to-image translation alone.

The limited availability of annotated data has driven us to pursue a self-supervised approach.
Data capture is much simpler without the requirement of ground truth, and furthermore the stereo pair does not need to be prealigned.

\subsection{Machine learning and self-supervised training}
\label{subsec:cnns}
FlowNet~\cite{Dosovitskiy2015a} from Dosovitskiy \etal~is a supervised end-to-end \ac{CNN} trained to estimate dense optical flow between RGB images.
It was succeeded by FlowNet 2.0~\cite{Ilg2017} from Ilg \etal, who improve the inference speed and accuracy by stacking sub-networks and having a more detailed training regime.
Although they are trained on synthetic datasets, they generalise well to real data such as KITTI~\cite{Geiger2013IJRR}.
Sun \etal~make a direct comparison to FlowNet2 with PWC\=/Net~\cite{Sun2018a}, another supervised network, which is smaller and easier to train.
Meister \etal~present UnFlow, a self-supervised approach~\cite{Meister2018}.
The authors use data losses between the warped image and the original, as well a consistency check between flow field directions.
Wang \etal~propose another self-supervised approach, UnDepthflow, which uses PWC\=/Net modules to isolate camera and scene motion~\cite{Wang2018a}.
UnFlow and UnDepthflow both achieve competitive accuracy on the KITTI dataset with other supervised networks.
Although the use-case is similar to ours, the above approaches cannot work with images of different spectra.

A major factor in the success of self-supervised approaches is the use of cycle-consistency~\cite{Godard2017,Godard2018a}.
In a cross-domain approach, Chen \etal~employ adversarial losses to for bidirectional domain transfer~\cite{Chen2020}.
This is similar to our approach in that the architecture consists of two halves which have mirrored operations.
However, Chen \etal~transfer between synthetic and real modalities (both RGB) as opposed to different spectra.
Aguilera \etal~learn a similarity measure for RGB to \ac{NIR} using siamese networks~\cite{Aguilera2016}.
Similarly, WILDCAT uses pseudo-siamese encoders to generate a shared latent space, to allow patch comparison between RGB and \ac{FIR}~\cite{Treible2019}.
Both of these require supervision to be trained, which limits their ability to generalise to different environments.

Generative approaches for RGB and \ac{NIR} use a cycleGAN to match across generated stereo pairs for both spectra~\cite{Liang2019a,Jeong2019}.
Performance of these image-to-image translation approaches are subject to spectral similarity.
Jeong \etal~show that for unsynchronized pairs, most approaches score lower on \ac{FIR} compared to \ac{NIR} when making feature-based comparisons~\cite{Jeong2019}.
We use synchronized pairs and assume all geometric differences are caused by the camera viewpoints.

\section{METHODOLOGY}
\label{sec:methodology}
Our goal is to enable dense correspondence between images with different spectra.
We estimate 2D flow fields between image pairs in both directions.
The data flow in our architecture is specifically designed to enable self-supervision.

\hfill\break\noindent\textbf{Data flow.}\quad
An overview of the losses and data flow is shown in \Cref{fig:overview}.
For simplicity, we only describe the operations for one half of the full forward pass, as the other is identical but inverted.
Modules \(\sigma_A\) and \(\sigma_B\) are flow estimation networks with the same architecture.
They have two encoder arms, one for each spectrum without sharing the weights.
We begin with two images, \(I_A\) and \(I_B\).
They describe the same scene but from different viewpoints and with different spectra.
Both are provided to modules \(\sigma_A\) and \(\sigma_B\).
Flow estimation module \(\sigma_B\) estimates how \(I_A\) should be warped to align with \(I_B\) and returns a 2D flow field, which we refer to as \({}^Af_B\).
Given \({}^Af_B\), the input image is warped using a differentiable sampling operation, producing \({}^AI_B\).
This shares \(I_A\)'s spectrum, but its structure should align with \(I_B\).
We formalize this as
\begin{equation}
	\label{eq:warp1}
	\begin{array}{>{\(}W{c}{0.45\linewidth}<{\)}>{\(}W{c}{0.45\linewidth}<{\)}}
		{}^Af_B = \sigma_B(I_A, I_B) &
		{}^Bf_A = \sigma_A(I_B, I_A)\\
		\\
		{}^AI_B = I_A \bigotimes {}^Af_B &
		{}^BI_A = I_B \bigotimes {}^Bf_A
		,
	\end{array}
\end{equation}
where we use the \(\bigotimes\) to represent a warping operation.
We supervise the flow estimates by warping back to the original input, and evaluating cycle-consistency.
\({}^AI_B\) is provided to other module \(\sigma_A\), whose task is to estimate the flow field, \({}^{AB}f_A\), to effectively undo the previous warping operation.
Similar to \cref{eq:warp1},
\begin{equation}
	\label{eq:warp2}
	\begin{array}{>{\(}W{c}{0.45\linewidth}<{\)}>{\(}W{c}{0.45\linewidth}<{\)}}
		{}^{AB}f_A = \sigma_A({}^AI_B, {}^BI_A) &
		{}^{BA}f_B = \sigma_B({}^BI_A, {}^AI_B)\\
		\\
		{}^{AB}I_A = {}^AI_B \bigotimes {}^{AB}f_A &
		{}^{BA}I_B = {}^BI_A \bigotimes {}^{BA}f_B
		.
	\end{array}
\end{equation}
\({}^{AB}I_A\) should now be a perfect reconstruction of \(I_A\) and may be compared directly.
Occlusions are handled implicitly by sampling interpolation, which is suitable for almost all cases.

\begin{figure}[t]\centering
	\includegraphics[width=\linewidth]{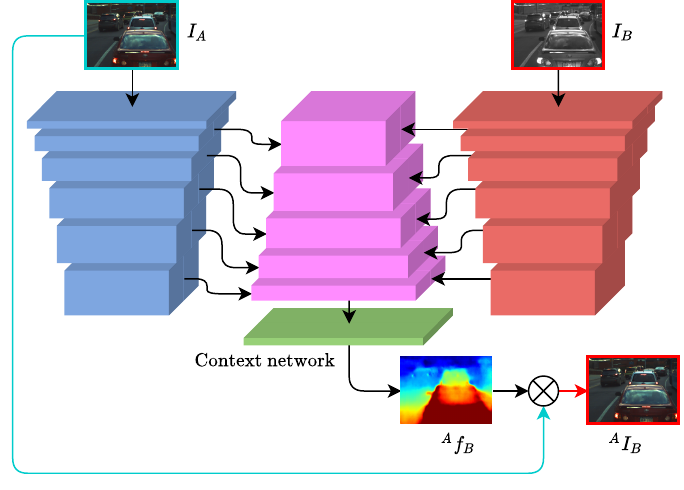}
	\caption{Flow estimation module \(\sigma\) architecture.
		Each red/blue block consists of 3 convolution layers, downsizing by a factor of 2.
		Purple blocks consist of warping, cost volume, optical flow estimation, and transpose convolution layers.
		In contrast to PWC\=/Net, we separate encoder weights (red \& blue).
	}
	\label{fig:module}
\end{figure}

\hfill\break\noindent\textbf{Flow estimation module.}\quad
The following relates to the module network structure in \cref{fig:module}.
The modules \(\sigma\) simultaneously encode images to 6-layer pyramids.
At each level, the features from one image are warped with the current optical flow estimate.
These warped features are combined with the features from the other image into a shared cost volume.
Each cost volume layer is fed through an optical flow estimator which passes the upscaled flow to the next layer, without sharing parameters between layers.
Basing on PWC\=/Net~\cite{Sun2018a}, we make the following enhancements:
\begin{reasons}
	\item We separate the encoder parameters so that each encoder may adapt to a different spectrum,
	\item We use suitable padding to overcome the restriction of tensor dimensions to multiples of \(2^n\) convolutional layers,
	\item We use strided transpose convolutions in the decoder in place of unweighted bilinear upsampling.
\end{reasons}

One forward pass of our full architecture uses each flow estimation module \(\sigma\) twice.
The tasks of each encoder in \(\sigma\) are very similar.
However, since our network is not targeted at any specific modality, the level of difference between aligned images is unknown.
Hence, sharing weights between each encoder may impair learning.
In \cref{fig:overview} we distinguish \(\sigma_A\) and \(\sigma_B\) as separate to indicate that the encoders do not share weights, but a shared decoder is used.
In each instance we swap the encoders at runtime to accommodate the input and target spectra, enabling parallel but separate learning.

\hfill\break\noindent\textbf{Cycle-consistency loss.}\quad
If an image can be warped away and then back, and remain equal to its original, then both flow fields agree.
The forward operation passes each image through both modules.
\(\sigma_A\) always warps from \(I_B\)'s spectrum to \(I_A\)'s, and \(\sigma_B\) from \(I_A\)'s to \(I_B\)'s.
By enforcing cycle-consistency, \(\sigma_A\) and \(\sigma_B\) must work together.
This encourages learned features to have a common representation between spectra.
Direct photometric image comparison also enables a high degree of reconstruction accuracy.
We introduce a simple mean-squared error over each pixel \(i\) for cycle-consistent supervision,
\begin{equation}
	\label{eq:cycleloss}
	\begin{array}{l}
		L_{C_1} = \sum\limits_{i=0}^{n-1}\left({{}^{AB}I_{A}}_i - {I_A}_i\right)^2\\
		L_{C_2} = \sum\limits_{i=0}^{n-1}\left({{}^{BA}I_{B}}_i - {I_B}_i\right)^2
		.
	\end{array}
\end{equation}
Each cycle-consistency loss backpropagates through both modules and maintains balance between them.

\hfill\break\noindent\textbf{Bidirectional flow-field loss.}\quad
Cycle-consistency is one way to measure if two flow fields cancel out, but it only applies to the serial flow fields, e.g.~\({}^Af_B\) with \({}^{AB}f_A\).
Parallel flow fields, such as \({}^Af_B\) and \({}^Bf_A\), should also cancel each other out.
We introduce a loss on the flow-fields directly to also supervise these cases in tandem with the cycle-consistency loss.
This is formalized as
\begin{equation}
	\label{eq:flowloss}
	\begin{array}{l}
		L_{B_1} =
		\left|
			{}^Af_B + ({}^Bf_A \bigotimes {}^Af_B)
		\right|
		,
	\end{array}
\end{equation}
where \(\bigotimes\) represents the warping function as an application of a flow field.
The similar operations for its inverse, and for the second stage (\({}^{AB}f_A\) and \({}^{BA}f_B\)), have been omitted for conciseness.

\hfill\break\noindent\textbf{Feature loss.}\quad
The trivial solution to minimising both the cycle-consistency loss \(L_C\) (\cref{eq:cycleloss}) and the bidirectional flow-field loss \(L_B\) (\cref{eq:flowloss}), is to generate flow fields which are entirely zero.
In this case, the loss would be zero and \({}^{AB}I_A\) would perfectly match the original \(I_A\).
We use a cross-spectral feature loss after the first warping stage to discourage this.

\({}^AI_B\) should align with \(I_B\), but their spectra are different.
To verify that the structure aligns while trying to ignoring spectrum differences, we use a perceptual loss~\cite{Johnson2016},
\begin{equation}
	\label{eq:edgeloss}
	\begin{array}{l}
		L_{F_1} = \frac{1}{CHW} || \phi({}^AI_B) - \phi(I_B) ||^2_2\\
		\\
		L_{F_2} = \frac{1}{CHW} || \phi({}^BI_A) - \phi(I_A) ||^2_2
		,
	\end{array}
\end{equation}
where \(\phi(x)\) are the layer activations of image \(x\), and \(C \times H \times W\) are the dimensions of that convolutional layer.
In the original work, Johnson \etal~defines features \(\phi\) at layer \texttt{relu3\_3} as suitable for matching content, with all layers suitable for style.
Roughly speaking, the deeper the layer at which latent features are extracted, the less the features will represent the original spectrum which is an important consideration for us.
However, due to the receptive field of the network, the deep layers can decrease precision.
It is important to consider the significant reduction in dimensions when trying to precisely align a scene.
For our problem, the best layer to use depends on the proximity of the spectra being compared.
For RGB-RGB correspondence, we found the activations of shallow layers suitable (\(\phi \in \texttt{relu2\_2}\)).

\hfill\break\noindent\textbf{Regularisation loss.}\quad
Correct estimated flow fields naturally warp some pixels out of the image boundaries.
In the absence of ground truth, gradients from the other losses cannot correct for pixels outside the image boundaries.
Flow fields may be generated with extreme values leading to an unrecoverable situation.
A low learning rate discourages this possibility but impairs the training, increasing the likelihood to get stuck in local minima.
To optimize the learning rate and achieve stability, we introduce a Huber loss \(L_R\) to act against unreasonable flow field magnitudes.
For flow fields \({}^Af_B\) and \({}^Bf_A\):
\begin{equation}
	\label{eq:regloss}
	\begin{array}{l}
		L_{R_{f}} =
		\begin{cases}
			\frac{1}{2} x_{f}^2 & \text{for } |x_{f}| < 1,\\
			|x_{f}| - \frac{1}{2} & \text{otherwise}
		\end{cases}
		\\
		\\
		\begin{array}{>{\(}W{c}{0.45\linewidth}<{\)}>{\(}W{c}{0.09\linewidth}<{\)}>{\(}W{c}{0.45\linewidth}<{\)}}
			x_f = \textit{max}(0, f - m) & \forall f \in \{{}^Af_B,{}^Bf_A\}
		,
		\end{array}
	\end{array}
\end{equation}
where \(m\) is defined as the maximum disparity in pixels which can be expected.
This is set to the disparity of the closest object in the dataset.
The regularisation loss only starts to affect flow field values above \(m\), which we typically set to 10\% of the image width.
The squared term at flow field, \(f\), lower than, \(x\), does not have a strong effect and permits a small amount of overshoot.
At higher values, the loss increases linearly to avoid exploding gradients.
In summary, pixels moved far from the image boundaries incur a high loss.
We only apply this on the flow fields produced at the first warping stage, as \({}^{AB}f_A\) and \({}^{BA}f_B\) are supervised by the cycle-consistency loss.

\hfill\break\noindent\textbf{Smoothing loss.}\quad
The flow fields themselves should be smooth except at depth discontinuities.
By enforcing uniformity in homogenous areas, the accuracy and visual quality of the warped images improves, thereby allowing a closer match for the cycle-consistency \(L_C\).
We introduce a term to penalize flow-field gradients which do not coincide with image gradients,
\begin{equation}
	\label{eq:smoothloss}
	\begin{array}{l}
		L_{S_1} = \left\lvert h({}^Af_B) * (1 - h(I_A)) \right\rvert\\
		L_{S_2} = \left\lvert h({}^Bf_A) * (1 - h(I_B)) \right\rvert\\
		L_{S_3} = \left\lvert h({}^{AB}f_A) * (1 - h({}^AI_B)) \right\rvert\\
		L_{S_4} = \left\lvert h({}^{BA}f_B) * (1 - h({}^BI_A)) \right\rvert
		,
	\end{array}
\end{equation}
where \(h(x)\) retrieves the Sobel gradient magnitude of \(x\), normalised between \num{0} and \num{1}.

\hfill\break\noindent\textbf{Combined loss function.}\quad
We define the overall loss as a balanced combination,
\begin{equation}
	\label{eq:totalloss}
	L_{\textit{total}} =
	\alpha{}L_C +
	\beta{}L_B +
	\gamma{}L_F +
	\delta{}L_R +
	\epsilon{}L_S
	,
\end{equation}
where \(\alpha\), \(\beta\), \(\gamma\), \(\delta\), \(\epsilon\) are training weights for each loss type.
For simplicity of notation, we refer to each loss type as the sum of its components, e.g.~\(L_C = L_{C_1} + L_{C_2}\).

\section{EXPERIMENTAL RESULTS}
\label{sec:results}
\noindent
\begin{table}[t]
	\setlength\belowcaptionskip{0pt}
	\caption{
		RGB-\ac{FIR} evaluation results for precision (\(Pr\)), recall (\(Re\)) and F1 score (\(F_1\)) on the VAP dataset~\cite{Palmero2016}.
		Scene 2 is omitted, as reported by other authors.
		As reference, the top three approaches (the highest performers for this dataset) estimate segmentation masks on registered images.
		The bottom three register ground truth masks using image data.
		The repeated baseline values represent the same evaluation.
		PWC\=/Net in parentheses was designed for fully-supervised RGB-RGB only.
	}
	\label{tab:fir}
	\setlength\tabcolsep{3pt}
	\begin{tabular}{@{\extracolsep{\stretch{1}}}lcrrrrrrr@{}}
		\toprule
		Method & Metric & \multicolumn{2}{c}{Scene 1} & \multicolumn{2}{c}{Scene 3} & \multicolumn{3}{c}{Overall}\\
		\cmidrule(lr){3-4}\cmidrule(lr){5-6}\cmidrule(ll){7-9}
		& & RGB & FIR & RGB & FIR & RGB & FIR & Mean\\
		\midrule
		\multirow{3}{*}{St-Charles~\cite{St-Charles2016}} & \(Pr\) & \color{g}0.820 & \color{g}0.755 & \color{g}0.716 & \color{g}0.514 & \color{g}0.768 & \color{g}0.635 & \color{g}0.701\\
		& \(Re\) & \color{g}0.810 & \color{g}\textbf{0.975} & \color{g}0.688 & \color{g}\textbf{0.969} & \color{g}0.749 & \color{g}\textbf{0.972} & \color{g}0.861\\
		& \(F_1\) & \color{g}0.815 & \color{g}0.851 & \color{g}0.702 & \color{g}0.672 & \color{g}0.758 & \color{g}0.762 & \color{g}0.760\\
		\midrule
		\multirow{3}{*}{GrabCut~\cite{Rother2004}} & \(Pr\) & \color{g}0.685 & \color{g}0.808 & \color{g}0.653 & \color{g}\textbf{0.847} & \color{g}0.669 & \color{g}\textbf{0.828} & \color{g}0.748\\
		& \(Re\) & \color{g}0.759 & \color{g}0.896 & \color{g}\textbf{0.929} & \color{g}0.916 & \color{g}0.844 & \color{g}0.906 & \color{g}0.875\\
		& \(F_1\) & \color{g}0.721 & \color{g}0.850 & \color{g}0.737 & \color{g}\textbf{0.880} & \color{g}0.744 & \color{g}\textbf{0.865} & \color{g}0.804\\
		\midrule
		\multirow{3}{*}{St-Charles~\cite{St-Charles2019}} & \(Pr\) & \color{g}\textbf{0.894} & \color{g}\textbf{0.860} & \color{g}\textbf{0.788} & \color{g}0.749 & \color{g}\textbf{0.841} & \color{g}0.804 & \color{g}\textbf{0.821}\\
		& \(Re\) & \color{g}\textbf{0.902} & \color{g}0.901 & \color{g}0.918 & \color{g}0.937 & \color{g}\textbf{0.910} & \color{g}0.919 & \color{g}\textbf{0.914}\\
		& \(F_1\) & \color{g}\textbf{0.898} & \color{g}\textbf{0.880} & \color{g}\textbf{0.848} & \color{g}0.833 & \color{g}\textbf{0.873} & \color{g}0.857 & \color{g}\textbf{0.866}\\
		\midrule
		\midrule
		\multirow{3}{*}{Baseline} & \(Pr\) & 0.536 & 0.536 & 0.559 & 0.559 & 0.548 & 0.548 & 0.548\\
		& \(Re\) & 0.525 & 0.525 & 0.535 & 0.535 & 0.530 & 0.530 & 0.530\\
		& \(F_1\) & 0.529 & 0.529 & 0.545 & 0.545 & 0.537 & 0.537 & 0.537\\
		\midrule
		\multirow{3}{*}{(PWC\=/Net)~\cite{Sun2018a}} & \(Pr\) & 0.448 & 0.337 & 0.693 & 0.129 & 0.571 & 0.233 & 0.402\\
		& \(Re\) & 0.497 & 0.279 & \textbf{0.729} & 0.091 & 0.613 & 0.185 & 0.399\\
		& \(F_1\) & 0.467 & 0.294 & \textbf{0.702} & 0.095 & 0.585 & 0.195 & 0.390\\
		\midrule
		\multirow{3}{*}{Ours} & \(Pr\) & \textbf{0.700} & \textbf{0.799} & \textbf{0.720} & \textbf{0.694} & \textbf{0.750} & \textbf{0.707} & \textbf{0.728}\\
		& \(Re\) & \textbf{0.569} & \textbf{0.768} & 0.518 & \textbf{0.779} & \textbf{0.669} & \textbf{0.649} & \textbf{0.659}\\
		& \(F_1\) & \textbf{0.622} & \textbf{0.781} & 0.595 & \textbf{0.734} & \textbf{0.702} & \textbf{0.665} & \textbf{0.683}\\
		\bottomrule
	\end{tabular}
	\vspace{-6mm}
\end{table}

In this section we evaluate our approach under different combinations of spectra.
We show that our approach is both capable of producing competitive state-of-the-art results, as well as generalising to more than a single fixed pair of spectra.
In order to demonstrate this, we evaluate three different cases:
{
\Crefname{section}{}{}
\begin{enumerate}
	\item \textbf{RGB-FIR}:
	We show that in addition to the previous two scenarios, our approach can recover flow fields and solve the correspondence problem between thermal and visible images.
	We evaluate the task of transferring annotations to the other spectrum.
	\item \textbf{RGB-NIR}:
	We show that without modifying the architecture, our approach can solve correspondence between the visible and \ac{NIR} spectra without ground truth supervision.
	We demonstrate its effectiveness with competitive evaluation scores on an automotive stereo dataset.
	\item \textbf{RGB-RGB}:
	We show that our self-supervised training approach is not limited to non-visible spectra, and it can still achieve competitive results when compared to other supervised state-of-the-art approaches.
\end{enumerate}
}
Each of these instances do not require a network modification, demonstrating the flexibility of our method.
For each, we train with a batch size of 8 and SGD optimizer, and optimize our hyperparameters with optuna~\cite{Akiba2019}.
For RGB-RGB, the learning rate was set to 4.3e-05, with losses weighted as
\(\alpha = \mathrm{\num{3.4e-1}}\), \(\beta = \mathrm{\num{3.6e-4}}\), \(\gamma = \mathrm{\num{6.7e-1}}\), \(\delta = \mathrm{\num{6.9e-2}}\), and \(\epsilon = \mathrm{\num{2.7e-1}}\).
Our approach is able to recover both the vertical and horizontal disparity, whereas many benchmarks are restricted to horizontal disparity.

\subsection{RGB-FIR evaluation}
\label{subsec:eval1}
\begin{table}[t]
	\setlength\belowcaptionskip{0pt}
	\caption{PittsStereo RGB-\ac{NIR} evaluation results.
	Score for each material category is the RMSE of disparity in pixels.
	PWC\=/Net in parentheses was designed for fully-supervised RGB-RGB only.}
	\label{tab:nir}
	\setlength\tabcolsep{2pt}
	\begin{tabular}{@{\extracolsep{\stretch{1}}}lrrrrrrrrrr@{}}
		\toprule
		Method & Bag & Cloth. & Com. & Glass & Glossy & Light & Skin & Veg. & Mean\\
		\midrule
		CMA~\cite{Chiu2011} & 4.63 & 6.42 & 1.60 & 2.55 & 3.86 & 5.17 & 3.39 & 4.42 & 4.00\\
		ANCC~\cite{YongSeokHeo2011} & 2.57 & 2.85 & 1.36 & 2.27 & 2.41 & 2.43 & 2.32 & 4.82 & 2.63\\
		DASC~\cite{Kim2015a} & 1.33 & \textbf{0.80} & 0.82 & 1.50 & 1.82 & 1.24 & 1.59 & 1.09 & 1.28\\
		\\
		Liang~\cite{Liang2019a} & \textbf{0.80} & 0.98 & 0.68 & \textbf{0.67} & \textbf{1.05} & \textbf{0.80} & 1.04 & \textbf{0.68} & \textbf{0.84}\\
		Zhi~\cite{Zhi2018} & 0.90 & 1.22 & \textbf{0.51} & 1.05 & 1.57 & 1.08 & \textbf{1.01} & 0.69 & 1.00\\
		(PWC\=/Net)~\cite{Sun2018a} & 4.88 & 5.67 & 11.20 & 6.06 & 3.33 & 3.92 & 7.00 & 13.5 & 6.95\\
		Ours & 0.91 & 0.90 & 0.64 & 1.18 & 1.49 & 1.00 & 1.47 & 1.10 & 1.08\\
		\bottomrule
	\end{tabular}
	\vspace{-2mm}
\end{table}

\begin{figure}[t]\centering
	\includegraphics[width=0.24\linewidth]{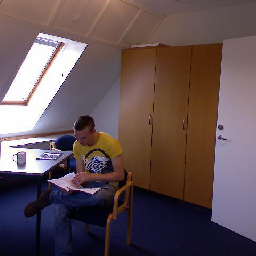}
	\hfill
	\includegraphics[width=0.24\linewidth]{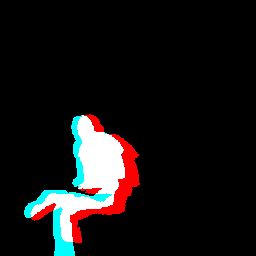}
	\includegraphics[width=0.24\linewidth]{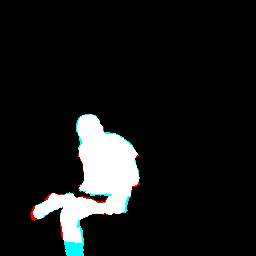}
	\includegraphics[width=0.24\linewidth]{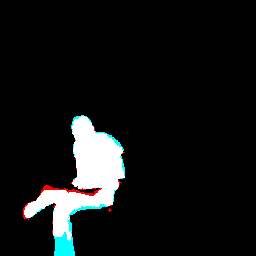}

	\includegraphics[width=0.24\linewidth]{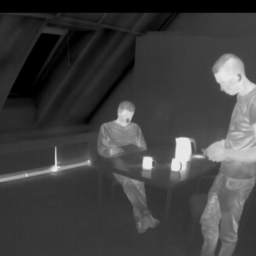}
	\hfill
	\includegraphics[width=0.24\linewidth]{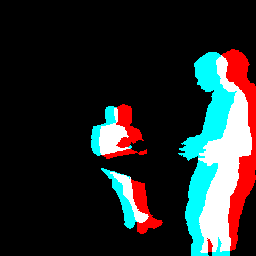}
	\includegraphics[width=0.24\linewidth]{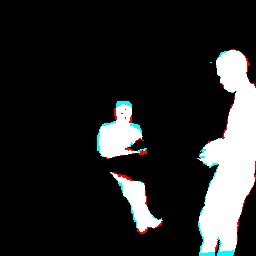}
	\includegraphics[width=0.24\linewidth]{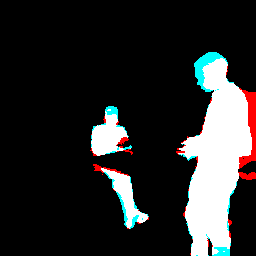}
	\caption{Qualitative results.
		\textbf{Column 1}: Example RGB and \ac{FIR} input images \(I_A\) and \(I_B\).
		\textbf{Column 2}: Original masks from \(\left[I_A, I_B\right]\), and used for baseline scores.
		\textbf{Column 3}: Warped RGB with \ac{FIR} masks \(\left[{}^AI_B, I_B\right]\).
		\textbf{Column 4}: Warped \ac{FIR} with RGB masks \(\left[{}^BI_A, I_A\right]\).}
	\label{fig:fir}
\end{figure}

We first evaluate on the VAP dataset~\cite{Palmero2016}, using the RGB and \ac{FIR} modalities only.
Ground truth correspondence is not provided.
We instead use human body segmentation as a proxy, by warping the ground truth segmentation mask from one spectrum to the other, and comparing against the other mask.
This process is significant because it provides an easier way to obtain annotations on challenging modalities.
A state-of-the-art RGB segmentation algorithm may be used to generate accurate masks which can then be densely registered to another viewpoint and spectrum.

\hfill\break\noindent\textbf{Training.}\quad
We break the left\(\rightarrow\)right RGB\(\rightarrow\)\ac{FIR} stereo assumption by horizontally flipping both images in each pair with a probability of \num{0.5} during training.
With data augmentation consistent between sample pairs, the static backgrounds may lead to overfitting to the camera distortion.
To prevent this we randomly crop pairs independently, i.e.~in a single pair, the RGB may be cropped to a slightly different region as the \ac{FIR}.
This introduces unpaired regions in each image, which are masked from the loss.
For each training sample, the network is forced to warp a region of the image, in any direction, to align and minimize the training losses.
This can only be achieved by learning how to match features across different spectra.

\hfill\break\noindent\textbf{Results.}\quad
The three approaches at the top of \cref{tab:fir} estimate the masks on registered images, whereas we use the provided masks with unregistered images.
This distinction is important when comparing scores.
When information is only present in one modality, a correct flow field will not give a perfect score, whereas segmentation on those images will reflect what is present.
In the qualitative comparison in \cref{fig:fir}, both RGB\(\rightarrow\)\ac{FIR} and \ac{FIR}\(\rightarrow\)RGB flow fields successfully align the masks (note that the masks are not seen by our network).
However, poor \ac{FIR} image quality around the borders results the bottom of its mask being cut off which harms our evaluation scores.

The `Baseline' row in \cref{tab:fir} represents the scores on unwarped, unregistered images (our evaluation's input).
The majority of the masks already overlap but are not aligned.
The decreased scores of PWC\=/Net~\cite{Sun2018a} shows the difficulty of the task, whereas ours improves the stereo registration.

{
\begin{figure}[h!]\centering
	\includegraphics[width=0.25\linewidth]{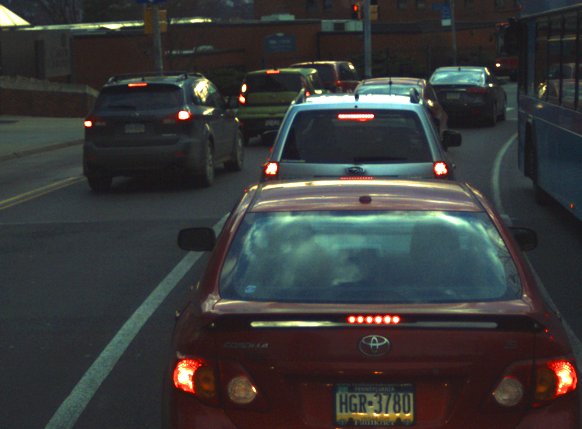}%
	\includegraphics[width=0.25\linewidth]{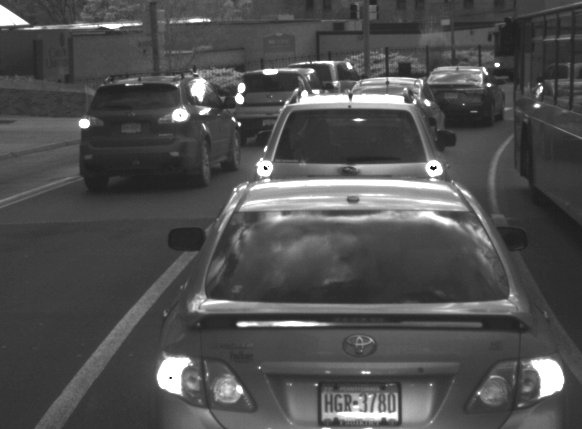}%
	\includegraphics[width=0.25\linewidth]{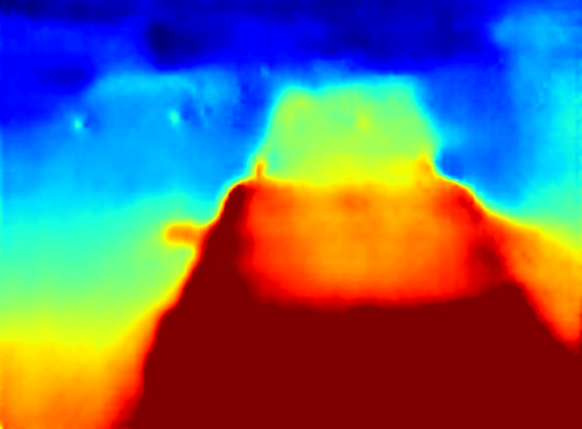}%
	\includegraphics[width=0.25\linewidth]{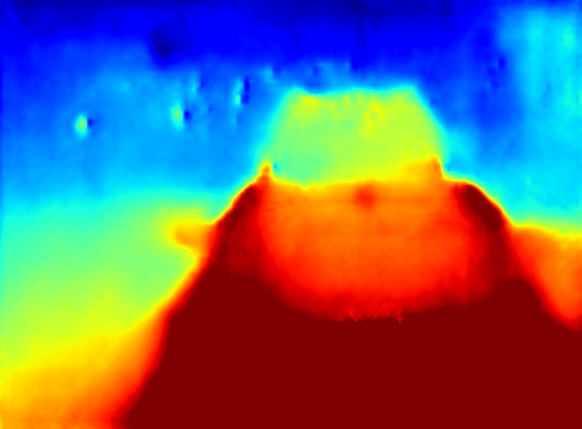}%

	\includegraphics[width=0.25\linewidth]{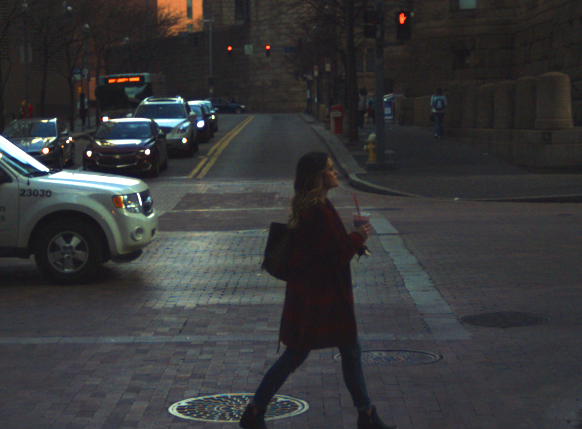}%
	\includegraphics[width=0.25\linewidth]{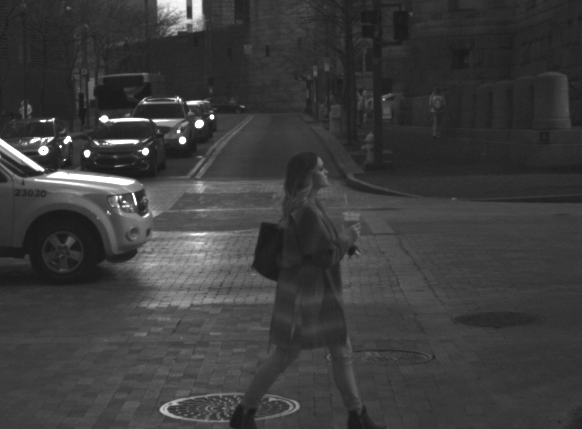}%
	\includegraphics[width=0.25\linewidth]{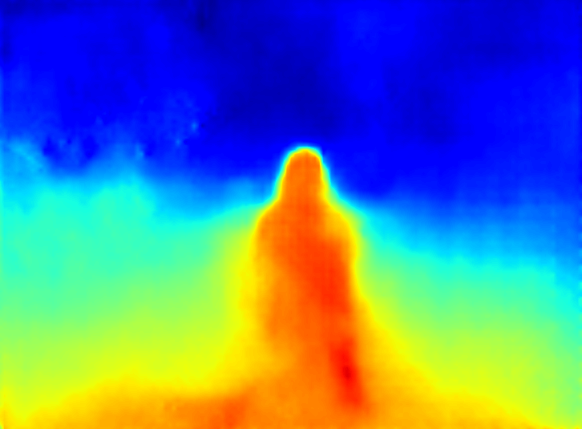}%
	\includegraphics[width=0.25\linewidth]{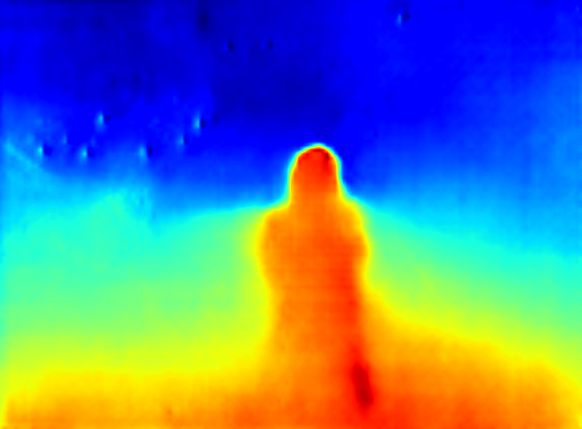}%

	\subcaptionbox{RGB
		\label{fig:FIG1}
	}
	[0.25\linewidth]{\includegraphics[width=\linewidth]{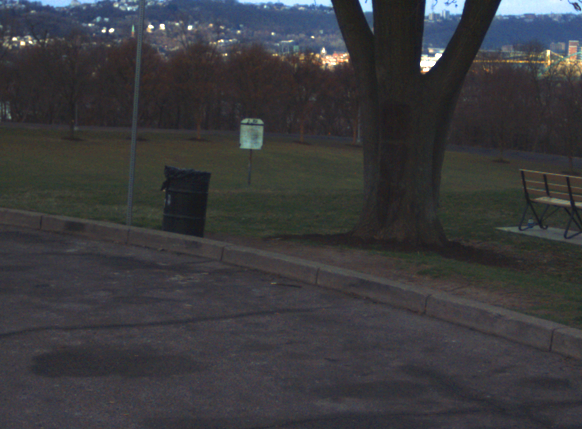}}%
	\subcaptionbox{\ac{NIR}
		\label{fig:FIG2}
	}
	[0.25\linewidth]{\includegraphics[width=\linewidth]{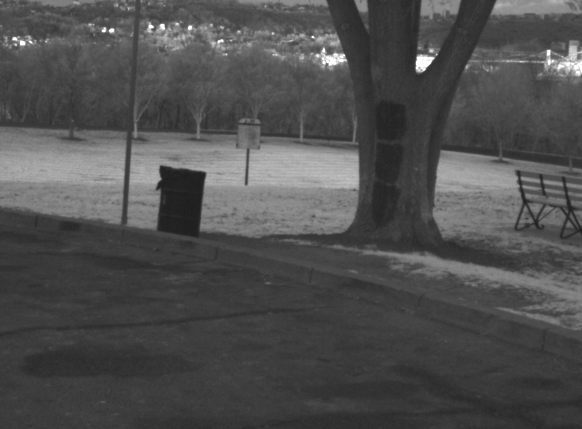}}%
	\subcaptionbox{RGB-\ac{NIR} flow
		\label{fig:FIG2}
	}
	[0.25\linewidth]{\includegraphics[width=\linewidth]{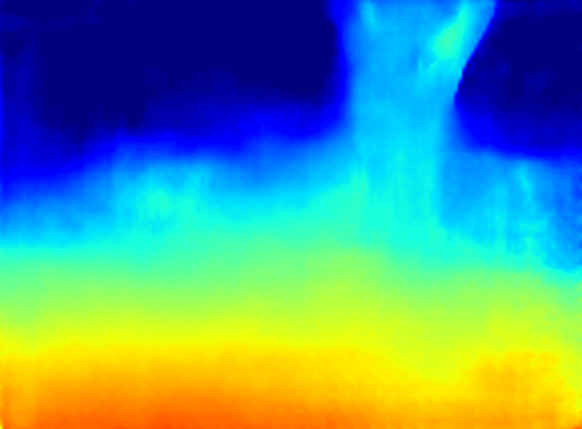}}%
	\subcaptionbox{\ac{NIR}-RGB flow
		\label{fig:FIG2}
	}
	[0.25\linewidth]{\includegraphics[width=\linewidth]{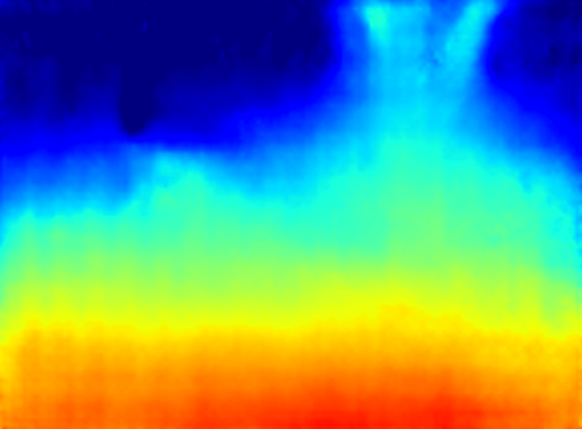}}%

	\setlength\abovecaptionskip{1.5\baselineskip}%
	\caption{Qualitative results on the RGB-\ac{NIR} evaluation dataset.
	Brightness and contrast increased for clarity.
	\textbf{Column 1}: Input RGB images \(I_A\).
	\textbf{Column 2}: Target \ac{NIR} images \(I_B\).
	\textbf{Column 3}: Flow fields \({}^Af_B\) \& \({}^Bf_A\) applied to input images.
	\textbf{Column 4}: Flow fields \({}^{AB}f_A\) \& \({}^{BA}f_B\) to return warped images to their input state.}
	\label{fig:nir}
\end{figure}
}

\subsection{RGB-NIR evaluation}
\label{subsec:eval2}
We also evaluate on the PittsStereo RGB-NIR dataset (automotive)~\cite{Zhi2018}.
The test set is annotated with a small number of point correspondences for each image pair, labelled with a material category including vegetation, glass, and lights.

\hfill\break\noindent\textbf{Training.}\quad
The dataset only exhibits horizontal disparity.
No 2D \ac{NIR} datasets could be found.
We do not restrict our network to 1D; it learns to ignore the vertical component.
We follow a similar training regime as for the RGB-\ac{FIR} (\cref{subsec:eval1}).
Since spectral bands are closer, we are able to augment the data by jittering the RGB values by 3\%.

\hfill\break\noindent\textbf{Results.}\quad
The RMSE error for each material type can be found in \Cref{tab:nir}.
Our approach clearly outperforms existing feature based methods.
The state-of-the-art deep learning approaches are specifically targeted at RGB-\ac{NIR}.
Our performance is comparable in spite of this, noting that errors are of the magnitude of a single pixel, accounting for \(<0.2\%\) of the image width.
\Cref{fig:nir} shows the flow fields estimated from both spectra and viewpoints agree, with clear object boundaries.

Both neural network approaches, from Zhi \etal~\cite{Zhi2018} and Liang \etal~\cite{Liang2019a} leverage image-to-image translation.
This is possible because the majority of materials in the \ac{NIR} spectrum closely resemble their grayscale counterparts in the visible image.
However, this is an inherent restriction preventing their transfer to other spectra.

\subsection{RGB-RGB evaluation}
\label{subsec:eval3}
We evaluate on the KITTI 2015 scene-flow dataset~\cite{Geiger2013IJRR}.
Despite the fact that all images are RGB, our design choices made to avoid direct left-right photometric image comparison do not cause an issue for same-spectrum correspondence estimation.
It should be noted that we do not make use of techniques specific to RGB-RGB correspondence estimation, nor do we estimate additional constraints such as vehicle egomotion.
We follow the same training regime as for the RGB-\ac{NIR} in \cref{subsec:eval2}.

\begin{figure}[t]\centering
	\includegraphics[width=0.5\linewidth]{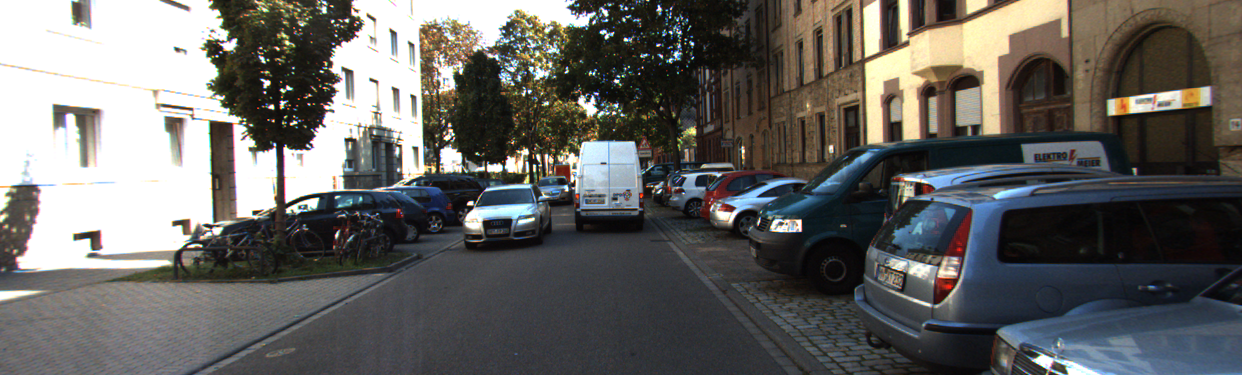}%
	\includegraphics[width=0.5\linewidth]{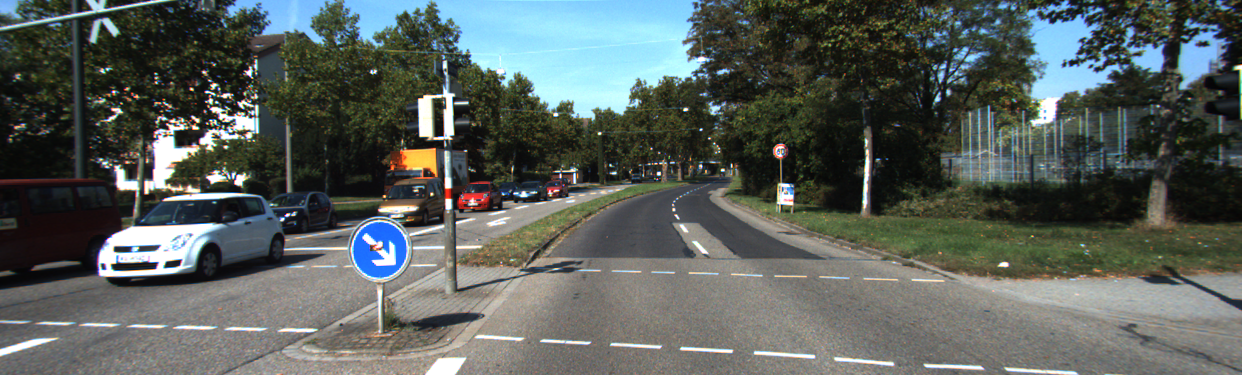}%

	\includegraphics[width=0.5\linewidth]{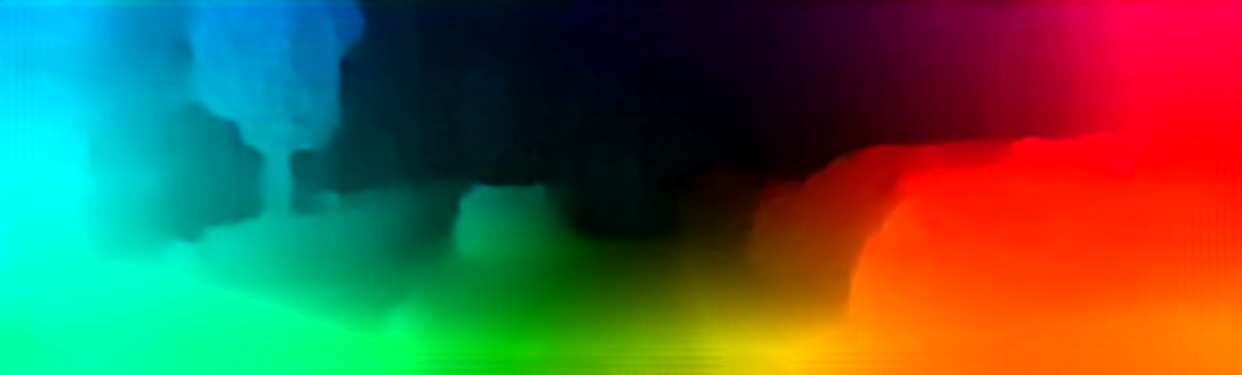}%
	\includegraphics[width=0.5\linewidth]{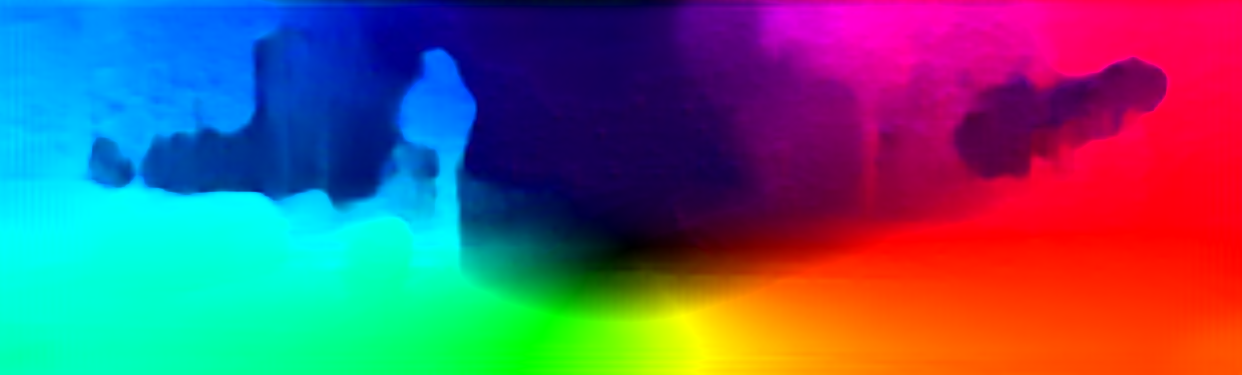}%

	\includegraphics[width=0.5\linewidth]{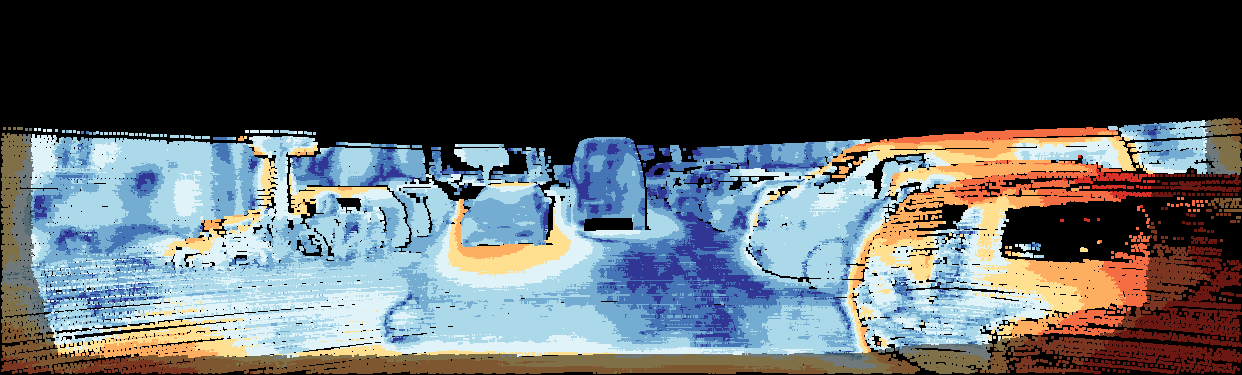}%
	\includegraphics[width=0.5\linewidth]{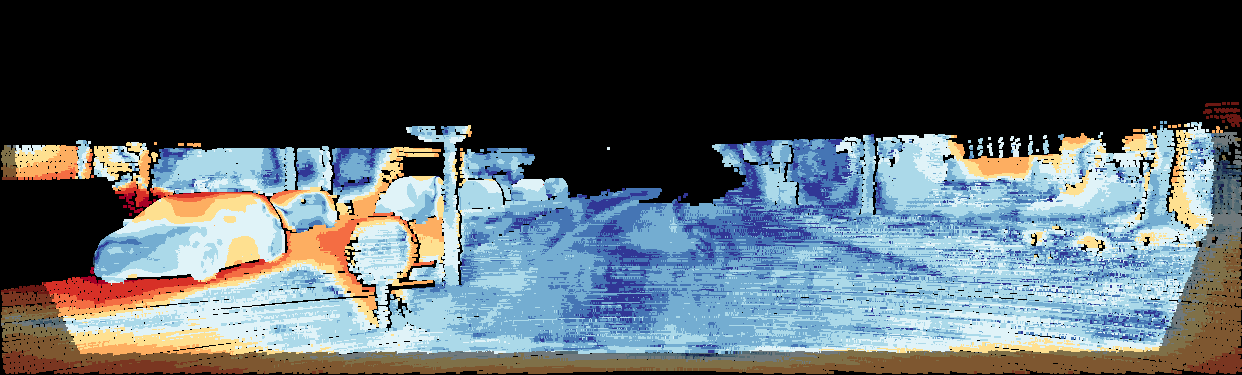}%
	\caption{KITTI 2015 evaluation (test set).
		\textbf{Row 1}: Input images \(I_A\).
		\textbf{Row 2}: Estimated flow fields \({}^Af_B\).
		\textbf{Row 3}: Error heatmaps.}
	\label{fig:rgb}
\end{figure}

\hfill\break\noindent\textbf{Results.}\quad
We compare against popular recent approaches and present quantitative results in \cref{tab:rgb}, including supervised approaches for context.
Compared with other recent self-supervised optical flow approaches, we achieve the lowest error on the test set.
This is even without using photometric loss or explicit occlusion reasoning.
None of the compared approaches are capable of cross-spectral matching.
Example qualitative flow fields and error heatmaps are shown in \cref{fig:rgb}.

\noindent
\begin{table}[t]
	\setlength\belowcaptionskip{0pt}
	\caption{KITTI optical flow evaluation 2015 results.
	`AEE' is the Average flow field End-Point Error in pixels.
	`Error' is the percentage of erroneous pixels, classed as having an end-point error of \(\geq\)3px and \(\geq\)5\%.}
	\label{tab:rgb}
	\begin{tabular*}{\linewidth}{@{\extracolsep{\stretch{1}}}lcrr@{}}
		\toprule
		Method & Self-supervised & AEE & Error\\
		\midrule
		FlowNet2~\cite{Ilg2017} & \xmark & 2.30 & 10.41\%\\
		UnFlow\=/CSS\=/ft~\cite{Meister2018} & \xmark & \textbf{1.86} & 11.11\%\\
		PWC\=/Net~\cite{Sun2018a} & \xmark & 2.16 & 9.60\%\\
		LiteFlowNet~\cite{Hui2018} & \xmark & 5.58 & 9.38\%\\
		LiteFlowNet2~\cite{Hui2019} & \xmark & 4.32 & \textbf{7.62}\%\\
		\midrule
		UnFlow\=/CSS~\cite{Meister2018} & \cmark & 8.10 & 23.30\%\\
		UnOS (FlowNet only)~\cite{Wang2019a,Wang2018a} & \cmark & 7.88 & 23.75\%\\
		Wang \etal~\cite{Wang2017} & \cmark & 8.88 & 31.20\%\\
		Janai \etal~\cite{Janai2018} & \cmark & 6.59 & 22.94\%\\
		DF\=/Net~\cite{zou2018dfnet} & \cmark & 8.98 & 25.70\%\\
		Ours & \cmark & \textbf{4.53} & \textbf{21.12}\%\\
		\bottomrule
	\end{tabular*}
	\vspace{-2mm}
\end{table}

\section{CONCLUSION}

In summary, we have presented a correspondence estimation method agnostic to the spectra that it is operating in.
We have shown that our approach tackles RGB-\acs{FIR}, RGB-\acs{NIR}, and RGB-RGB.
This is all while achieving results comparable to state-of-the-art methods, many of which that have been fine-tuned to their specific modality.
More generally we have demonstrated that, at a fundamental level, cross-spectral correspondence estimation is a problem that can be solved in a generic way.
Additionally, our approach demonstrates that this can all be done in a self-supervised manner.
This enables correspondence estimation algorithms to be trained in domains that have very little training data and virtually no annotations.
With this, we hope this that the interest in the cross-spectral domain continues to grow, to expand the currently limited variety of datasets and algorithms.
\balance

\bibliographystyle{ieeetran}
\bibliography{bibliography}

\begin{thebibliography}{10}
\providecommand{\url}[1]{#1}
\csname url@rmstyle\endcsname
\providecommand{\newblock}{\relax}
\providecommand{\bibinfo}[2]{#2}
\providecommand\BIBentrySTDinterwordspacing{\spaceskip=0pt\relax}
\providecommand\BIBentryALTinterwordstretchfactor{4}
\providecommand\BIBentryALTinterwordspacing{\spaceskip=\fontdimen2\font plus
\BIBentryALTinterwordstretchfactor\fontdimen3\font minus
  \fontdimen4\font\relax}
\providecommand\BIBforeignlanguage[2]{{%
\expandafter\ifx\csname l@#1\endcsname\relax
\typeout{** WARNING: IEEEtran.bst: No hyphenation pattern has been}%
\typeout{** loaded for the language `#1'. Using the pattern for}%
\typeout{** the default language instead.}%
\else
\language=\csname l@#1\endcsname
\fi
#2}}

\bibitem{Hartley2004}
R.~I. Hartley and A.~Zisserman, \emph{Multiple View Geometry in Computer
  Vision}, 2nd~ed.\hskip 1em plus 0.5em minus 0.4em\relax Cambridge University
  Press, ISBN: 0521540518, 2004.

\bibitem{Brady1995}
J.~M. Brady, ``{ASSET-2: Real-Time Motion Segmentation and Shape Tracking},''
  \emph{IEEE Transactions on Pattern Analysis and Machine Intelligence},
  vol.~17, no.~8, pp. 814--820, 1995.

\bibitem{Lowe2004}
D.~G. Lowe, ``{Distinctive image features from scale-invariant keypoints},''
  \emph{International Journal of Computer Vision}, vol.~60, no.~2, pp. 91--110,
  2004.

\bibitem{Lucas1981}
B.~D. Lucas and T.~Kanade, ``{An Iterative Image Registration Technique with an
  Application to Stereo Vision},'' \emph{DARPA Image Understanding Workshop},
  pp. 121--130, 1981.

\bibitem{Besl1992a}
P.~Besl and N.~McKay, ``{A Method for Registration of 3-D Shapes},'' \emph{IEEE
  Trans. Pattern Anal. Mach. Intell.}, vol.~14, no.~2, pp. 239--256, 1992.

\bibitem{Liu2015}
C.~Liu, J.~Yuen, and A.~Torralba, ``{Sift flow: Dense correspondence across
  scenes and its applications},'' \emph{Dense Image Correspondences for
  Computer Vision}, pp. 15--49, 2015.

\bibitem{Tola2010}
E.~Tola, V.~Lepetit, and P.~Fua, ``{DAISY: An efficient dense descriptor
  applied to wide-baseline stereo},'' \emph{IEEE Transactions on Pattern
  Analysis and Machine Intelligence}, vol.~32, no.~5, pp. 815--830, 2010.

\bibitem{Dosovitskiy2015a}
A.~Dosovitskiy, P.~Fischery, E.~Ilg, P.~Hausser, C.~Hazirbas, V.~Golkov,
  P.~V.~D. Smagt, D.~Cremers, and T.~Brox, ``{FlowNet: Learning optical flow
  with convolutional networks},'' \emph{Proceedings of the IEEE International
  Conference on Computer Vision}, vol. 2015 Inter, pp. 2758--2766, 2015.

\bibitem{Ilg2017}
E.~Ilg, N.~Mayer, T.~Saikia, M.~Keuper, A.~Dosovitskiy, and T.~Brox, ``{FlowNet
  2.0: Evolution of optical flow estimation with deep networks},''
  \emph{Proceedings - 30th IEEE Conference on Computer Vision and Pattern
  Recognition, CVPR 2017}, vol. 2017-Janua, pp. 1647--1655, 2017.

\bibitem{Sun2018a}
D.~Sun, X.~Yang, M.~Y. Liu, and J.~Kautz, ``{PWC-Net: CNNs for Optical Flow
  Using Pyramid, Warping, and Cost Volume},'' \emph{Proc. IEEE Comput. Soc.
  Conf. Comput. Vis. Pattern Recognit.}, vol.~D, pp. 8934--8943, 2018.

\bibitem{Meister2018}
S.~Meister, J.~Hur, and S.~Roth, ``{UnFlow: Unsupervised learning of optical
  flow with a bidirectional census loss},'' \emph{32nd AAAI Conference on
  Artificial Intelligence, AAAI 2018}, pp. 7251--7259, 2018.

\bibitem{Liu2019}
X.~Liu, Y.~Ai, B.~Tian, and D.~Cao, ``{Robust and Fast Registration of Infrared
  and Visible Images for Electro-Optical Pod},'' \emph{IEEE Transactions on
  Industrial Electronics}, vol.~66, no.~2, pp. 1335--1344, 2019.

\bibitem{Hwang2015}
S.~Hwang, J.~Park, N.~Kim, Y.~Choi, and I.~S. Kweon, ``{Multispectral
  pedestrian detection: Benchmark dataset and baseline},'' \emph{Proceedings of
  the IEEE Computer Society Conference on Computer Vision and Pattern
  Recognition}, vol. 07-12-June, pp. 1037--1045, 2015.

\bibitem{Okazawa2019}
A.~Okazawa, T.~Takahata, and T.~Harada, ``{Simultaneous transparent and
  non-transparent object segmentation with multispectral scenes},'' in
  \emph{2019 IEEE/RSJ International Conference on Intelligent Robots and
  Systems (IROS2019)}, 2019, pp. 4977--4984.

\bibitem{Gebhardt2019}
E.~Gebhardt and M.~Wolf, ``{CAMEL Dataset for Visual and Thermal Infrared
  Multiple Object Detection and Tracking},'' \emph{Proceedings of AVSS 2018 -
  2018 15th IEEE International Conference on Advanced Video and Signal-Based
  Surveillance}, pp. 1--6, 2019.

\bibitem{Davis2005}
J.~W. Davis and M.~A. Keck, ``{A Two-Stage Template Approach to Person
  Detection in Thermal Imagery},'' in \emph{Proc. Workshop on Applications of
  Computer Vision}, 2005.

\bibitem{Zhang2015d}
M.~M. Zhang, J.~Choi, K.~Daniilidis, M.~T. Wolf, and C.~Kanan, ``{VAIS: A
  dataset for recognizing maritime imagery in the visible and infrared
  spectrums},'' \emph{IEEE Computer Society Conference on Computer Vision and
  Pattern Recognition Workshops}, vol. 2015-Octob, pp. 10--16, 2015.

\bibitem{Li2016a}
C.~Li, H.~Cheng, S.~Hu, X.~Liu, J.~Tang, and L.~Lin, ``{Learning Collaborative
  Sparse Representation for Grayscale-Thermal Tracking},'' \emph{IEEE
  Transactions on Image Processing}, vol.~25, no.~12, pp. 5743--5756, 2016.

\bibitem{Li2018}
C.~Li, X.~Liang, Y.~Lu, N.~Zhao, and J.~Tang, ``{RGB-T Object Tracking:
  Benchmark and Baseline},'' pp. 1--13, 2018.

\bibitem{Ricaurte2014}
P.~Ricaurte, C.~Chil{\'{a}}n, C.~A. Aguilera-Carrasco, B.~X. Vintimilla, and
  A.~D. Sappa, ``{Feature point descriptors: Infrared and visible spectra},''
  \emph{Sensors (Switzerland)}, vol.~14, no.~2, pp. 3690--3701, 2014.

\bibitem{Brown2011}
M.~Brown and S.~S{\"{u}}sstrunk, ``{Multispectral SIFT for scene category
  recognition},'' in \emph{Proceedings of the IEEE Computer Society Conference
  on Computer Vision and Pattern Recognition}, 2011, pp. 177--184.

\bibitem{Saleem2014}
S.~Saleem and R.~Sablatnig, ``{A robust SIFT descriptor for multispectral
  images},'' \emph{IEEE Signal Processing Letters}, vol.~21, no.~4, pp.
  400--403, 2014.

\bibitem{Kim2015a}
S.~Kim, D.~Min, B.~Ham, S.~Ryu, M.~N. Do, and K.~Sohn, ``{DASC: Dense adaptive
  self-correlation descriptor for multi-modal and multi-spectral
  correspondence},'' \emph{Proceedings of the IEEE Computer Society Conference
  on Computer Vision and Pattern Recognition}, vol. 07-12-June, pp. 2103--2112,
  2015.

\bibitem{Hajebi2008}
K.~Hajebi and J.~S. Zelek, ``{Structure from infrared stereo images},''
  \emph{Proceedings of the 5th Canadian Conference on Computer and Robot
  Vision, CRV 2008}, pp. 105--112, 2008.

\bibitem{Mouats2015}
T.~Mouats, N.~Aouf, L.~Chermak, and M.~A. Richardson, ``{Thermal stereo
  odometry for UAVs},'' \emph{IEEE Sensors Journal}, vol.~15, no.~11, pp.
  6335--6347, 2015.

\bibitem{Li2017}
Y.~Li and R.~L. Stevenson, ``{Multimodal Image Registration With Line Segments
  by Selective Search},'' \emph{IEEE Transactions on Cybernetics}, vol.~47,
  no.~5, pp. 1285--1298, 2017.

\bibitem{Viola1997}
P.~Viola and W.~M. Wells, ``{Alignment by Maximization of Mutual
  Information},'' \emph{International Journal of Computer Vision}, vol.~24,
  no.~2, pp. 137--154, 1997.

\bibitem{Belghazi2018}
M.~I. Belghazi, A.~Baratin, S.~Rajeswar, S.~Ozair, Y.~Bengio, A.~Courville, and
  R.~D. Hjelm, ``{Mutual Information Neural Estimation},'' \emph{35th
  International Conference on Machine Learning, ICML 2018}, vol.~2, pp.
  864--873, 2018.

\bibitem{Choe2018}
G.~Choe, S.~H. Kim, S.~Im, J.~Y. Lee, S.~G. Narasimhan, and I.~S. Kweon,
  ``{RANUS: RGB and NIR urban scene dataset for deep scene parsing},''
  \emph{IEEE Robotics and Automation Letters}, vol.~3, no.~3, pp. 1808--1815,
  2018.

\bibitem{Bilodeau2014}
G.~A. Bilodeau, A.~Torabi, P.~L. St-Charles, and D.~Riahi, ``{Thermal-visible
  registration of human silhouettes: A similarity measure performance
  evaluation},'' \emph{Infrared Physics and Technology}, vol.~64, pp. 79--86,
  2014.

\bibitem{Zhi2018}
T.~Zhi, B.~R. Pires, M.~Hebert, and S.~G. Narasimhan, ``{Deep Material-Aware
  Cross-Spectral Stereo Matching},'' in \emph{IEEE Conf. Comput. Vis. Pattern
  Recognit.}, 2018.

\bibitem{Palmero2016}
C.~Palmero, A.~Clap{\'{e}}s, C.~Bahnsen, A.~M{\o}gelmose, T.~B. Moeslund, and
  S.~Escalera, ``{Multi-modal RGB–Depth–Thermal Human Body Segmentation},''
  \emph{Int. J. Comput. Vis.}, vol. 118, pp. 217--239, 2016.

\bibitem{Treible2017}
W.~Treible, P.~Saponaro, S.~Sorensen, A.~Kolagunda, M.~O'Neal, B.~Phelan,
  K.~Sherbondy, and C.~Kambhamettu, ``{CATS: A color and thermal stereo
  benchmark},'' in \emph{Conference on Computer Vision and Pattern Recognition
  (CVPR)}, 2017, pp. 2961--2969.

\bibitem{Li2019}
C.~Li, W.~Xia, Y.~Yan, B.~Luo, and J.~Tang, ``{Segmenting Objects in Day and
  Night: Edge-Conditioned CNN for Thermal Image Semantic Segmentation},''
  vol.~1, pp. 1--12, 2019.

\bibitem{Geiger2013IJRR}
A.~Geiger, P.~Lenz, C.~Stiller, and R.~Urtasun, ``{Vision meets robotics: The
  KITTI dataset},'' \emph{International Journal of Robotics Research}, vol.~32,
  no.~11, pp. 1231--1237, 2013.

\bibitem{Wang2018a}
Y.~Wang, Z.~Yang, P.~Wang, Y.~Yang, C.~Luo, and W.~Xu, ``{Joint Unsupervised
  Learning of Optical Flow and Depth by Watching Stereo Videos},'' oct 2018.

\bibitem{Godard2017}
C.~Godard, O.~{Mac Aodha}, and G.~J. Brostow, ``{Unsupervised monocular depth
  estimation with left-right consistency},'' \emph{Proceedings - 30th IEEE
  Conference on Computer Vision and Pattern Recognition, CVPR 2017}, vol.
  2017-Janua, pp. 6602--6611, 2017.

\bibitem{Godard2018a}
C.~Godard, O.~{Mac Aodha}, M.~Firman, and G.~Brostow, ``{Digging Into
  Self-Supervised Monocular Depth Estimation},'' no.~1, 2018.

\bibitem{Chen2020}
Y.-C. Chen, Y.-Y. Lin, M.-H. Yang, and J.-B. Huang, ``{CrDoCo: Pixel-Level
  Domain Transfer With Cross-Domain Consistency},'' pp. 1791--1800, 2020.

\bibitem{Aguilera2016}
C.~A. Aguilera, F.~J. Aguilera, A.~D. Sappa, and R.~Toledo, ``{Learning
  Cross-Spectral Similarity Measures with Deep Convolutional Neural
  Networks},'' \emph{IEEE Computer Society Conference on Computer Vision and
  Pattern Recognition Workshops}, pp. 267--275, 2016.

\bibitem{Treible2019}
W.~Treible, P.~Saponaro, Y.~Liu, A.~{Das Gupta}, V.~Veerendraveer, S.~Sorensen,
  and C.~Kambhamettu, ``{CATS 2: Color And Thermal Stereo Scenes with Semantic
  Labels},'' in \emph{CVPR Workshops}, 2019, pp. 0--4.

\bibitem{Liang2019a}
M.~Liang, X.~Guo, H.~Li, X.~Wang, and Y.~Song, ``{Unsupervised Cross-spectral
  Stereo Matching by Learning to Synthesize},'' 2019.

\bibitem{Jeong2019}
S.~Jeong, S.~Kim, K.~Park, and K.~Sohn, ``{Learning to Find Unpaired
  Cross-Spectral Correspondences},'' \emph{IEEE Transactions on Image
  Processing}, vol.~28, no.~11, pp. 5394--5406, 2019.

\bibitem{Johnson2016}
J.~Johnson, A.~Alahi, and L.~Fei-Fei, ``{Perceptual losses for real-time style
  transfer and super-resolution},'' \emph{Lect. Notes Comput. Sci. (including
  Subser. Lect. Notes Artif. Intell. Lect. Notes Bioinformatics)}, vol. 9906
  LNCS, pp. 694--711, 2016.

\bibitem{St-Charles2016}
P.~L. St-Charles, G.~A. Bilodeau, and R.~Bergevin, ``{Universal Background
  Subtraction Using Word Consensus Models},'' \emph{IEEE Trans. Image
  Process.}, vol.~25, no.~10, pp. 4768--4781, 2016.

\bibitem{Rother2004}
C.~Rother, V.~Kolmogorov, and A.~Blake, ``{"GrabCut"- Interactive foreground
  extraction using iterated graph cuts},'' \emph{ACM Trans. Graph. -- Proc. ACM
  SIGGRAPH 2004}, vol.~23, no.~3, p. 309, 2004.

\bibitem{St-Charles2019}
P.~L. St-Charles, G.~A. Bilodeau, and R.~Bergevin, ``{Online Mutual Foreground
  Segmentation for Multispectral Stereo Videos},'' \emph{International Journal
  of Computer Vision}, vol. 127, no.~8, pp. 1044--1062, 2019.

\bibitem{Akiba2019}
T.~Akiba, S.~Sano, T.~Yanase, T.~Ohta, and M.~Koyama, ``{Optuna: A
  next-generation hyperparameter optimization framework},'' \emph{Proceedings
  of the 25rd {ACM} {SIGKDD} International Conference on Knowledge Discovery
  and Data Mining}, pp. 2623--2631, 2019.

\bibitem{Chiu2011}
W.~C. Chiu, U.~Blanke, and M.~Fritz, ``{Improving the kinect by cross-modal
  stereo},'' \emph{BMVC 2011 - Proc. Br. Mach. Vis. Conf. 2011}, pp. 1--10,
  2011.

\bibitem{YongSeokHeo2011}
{Yong Seok Heo}, {Kyong Mu Lee}, and {Sang Uk Lee}, ``{Robust Stereo Matching
  Using Adaptive Normalized Cross-Correlation},'' \emph{IEEE Trans. Pattern
  Anal. Mach. Intell.}, vol.~33, no.~4, pp. 807--822, apr 2011.

\bibitem{Hui2018}
T.~W. Hui, X.~Tang, and C.~C. Loy, ``{LiteFlowNet: A Lightweight Convolutional
  Neural Network for Optical Flow Estimation},'' \emph{Proc. IEEE Comput. Soc.
  Conf. Comput. Vis. Pattern Recognit.}, pp. 8981--8989, 2018.

\bibitem{Hui2019}
T.-W. Hui, X.~Tang, and C.~C. Loy, ``{A Lightweight Optical Flow CNN -
  Revisiting Data Fidelity and Regularization},'' pp. 1--15, 2019.

\bibitem{Wang2019a}
Y.~Wang, P.~Wang, Z.~Yang, C.~Luo, Y.~Yang, and W.~Xu, ``{UnOS: Unified
  unsupervised optical-flow and stereo-depth estimation by watching videos},''
  \emph{Proc. IEEE Comput. Soc. Conf. Comput. Vis. Pattern Recognit.}, vol.
  2019-June, pp. 8063--8073, 2019.

\bibitem{Wang2017}
Y.~Wang, Y.~Yang, Z.~Yang, L.~Zhao, P.~Wang, and W.~Xu, ``{Occlusion Aware
  Unsupervised Learning of Optical Flow},'' nov 2017.

\bibitem{Janai2018}
J.~Janai, F.~G{\"{u}}ney, A.~Ranjan, M.~Black, and A.~Geiger, ``{Unsupervised
  Learning of Multi-Frame Optical Flow with Occlusions},'' \emph{Lect. Notes
  Comput. Sci. (including Subser. Lect. Notes Artif. Intell. Lect. Notes
  Bioinformatics)}, vol. 11220 LNCS, pp. 713--731, 2018.

\bibitem{zou2018dfnet}
Y.~Zou, Z.~Luo, J.-b. Huang, and I.~Separate, ``{DF-Net: Unsupervised Joint
  Learning of Depth and Flow using Cross-Task Consistency},'' \emph{Eur. Conf.
  Comput. Vis.}, 2018.

\end{thebibliography}
\end{document}